\documentclass[runningheads]{llncs}
\usepackage{graphicx}

\usepackage{tikz}
\usepackage{comment}
\usepackage{amsmath,amssymb} 
\usepackage{color}

\usepackage[accsupp]{axessibility}  

\definecolor{myGreen}{RGB}{0,114,0}
\definecolor{Gray}{gray}{0.95}
\definecolor{turquoise}{RGB}{64,224,208}
\definecolor{pink1}{RGB}{247,146,248}
\definecolor{cyan1}{RGB}{145,253,254}
\usepackage[pagebackref,breaklinks,colorlinks]{hyperref}
\usepackage{multirow}
\usepackage{booktabs}
\usepackage{bm}
\usepackage{mathtools}
\usepackage{xcolor,colortbl}
\usepackage{hhline}

\begin{document}
	\pagestyle{headings}
	\mainmatter
	\def\ECCVSubNumber{7090}  
	
	\title{Learning Omnidirectional Flow in 360\texorpdfstring{$^\circ$}{} Video via Siamese Representation} 

		\titlerunning{Learning Omnidirectional Flow in 360\texorpdfstring{$^\circ$}{} Video via Siamese Representation}
		%
		\author{Keshav Bhandari\inst{1} \and
			Bin Duan\inst{2} \and
			Gaowen Liu\inst{3} \and
			Hugo Latapie\inst{3} \and 
			Ziliang Zong\inst{1} \and
			Yan Yan\inst{2}\thanks{Corresponding author.}
		}
		\authorrunning{K. Bhandari et al.}
		%
		\institute{$^1$Texas State University \, $^2$Illinois Institute of Technology \, $^3$Cisco Research
		}
	\maketitle
	\begin{abstract}
\label{sec:abstract}
Optical flow estimation in omnidirectional videos faces two significant issues: the lack of benchmark datasets and the challenge of adapting perspective video-based methods to accommodate the omnidirectional nature. This paper proposes the first perceptually natural-synthetic omnidirectional benchmark dataset with a 360\texorpdfstring{$^\circ$}{} field of view, FLOW360, with 40 different videos and 4,000 video frames. We conduct comprehensive characteristic analysis and comparisons between our dataset and existing optical flow datasets, which manifest perceptual realism, uniqueness, and diversity. To accommodate the omnidirectional nature, we present a novel Siamese representation Learning framework for Omnidirectional Flow (SLOF). We train our network in a contrastive manner with a hybrid loss function that combines contrastive loss and optical flow loss. Extensive experiments verify the proposed framework's effectiveness and show up to 40\% performance improvement over the state-of-the-art approaches. Our FLOW360 dataset and code are available at \url{https://siamlof.github.io/}.

\end{abstract}

	\section{Introduction}
\label{sec:intro}
\begin{figure}[!t]
    \centering
    \includegraphics[width=0.99\textwidth]{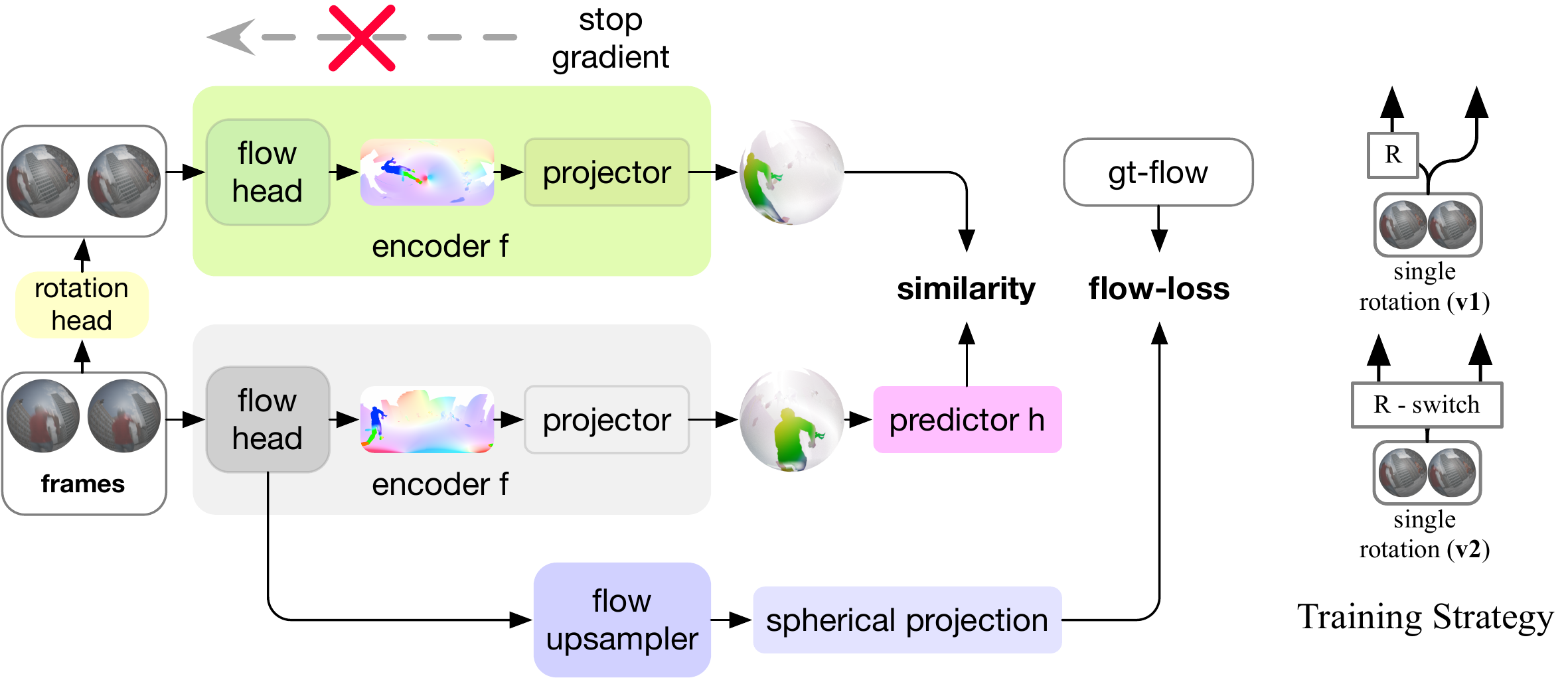}
    \caption{\textbf{Siamese Representation Learning for Omnidirectional Flow (SLOF)}. Pairs of frame sequence (w/ and w/o random rotation) are passed as inputs to encoder \textit{f} (RAFT as a flow head backbone and a standard convolutional projector layer). A predictor layer \textit{h} is an MLP layer. The entire framework is trained by fusing the pretraining and fine-tuning stage to combine the similarity and flow-loss in a single stage. The model maximizes the similarity between latent representations of flow information from two streams and minimizes the flow loss. \textbf{Training Strategy (right):} Here two different arrows$(left,right)$ represent siamese streams or input pathways to our model. \textbf{v1} and \textbf{v2} (either of the stream is subjected to rotational augmentation) are similar strategies achieving overall better performance.}
    \label{fig:architecture}
    \vspace{-0.2in}
\end{figure}

Optical flow estimation, as a fundamental problem in computer vision, has been studied over decades by early works~\cite{lucas,horn} dated back to 80\textit{s}. Before the era of modern deep learning, traditional optical flow estimation methods relied on hand-crafted features based optimizations~\cite{pixel1,pixel2,pixel3}, energy-based optimizations~\cite{energy1,energy2,energy3} and variational approaches~\cite{feature1,variational1,feature2}. Although deep learning-based approaches~\cite{gma,raft,maskflownet,pwc,flownet2,liteflownet} have shown great advantages over these classical approaches, most of them are specially tailored for perspective videos. The availability of perspective optical flow datasets~\cite{sintel,kitti,kitti2012,kitti2015,middlebury} heavily supports the advancement of these modern deep learning-based approaches. The optical flow datasets are difficult to obtain and requires the generation of naturalistic synthetic dataset like Sintel~\cite{sintel}. As these datasets mark the foundation for optical flow estimation research, the availability of reliable omnidirectional datasets is equally important to advance the omnidirectional flow estimation research. The need for the datasets brings up the first challenge: there is no such reliable (perceptually natural and complex) 360$^\circ$ or omnidirectional video dataset in the literature collected for omnidirectional optical flow estimation. Another challenge of omnidirectional optical flow estimation is that current perspective video-based deep networks fail to accommodate the nature of 360$^\circ$ videos. These perspective optical flow estimation methods inevitably require fine-tuning due to the presence of radial distortion~\cite{radial} on 360$^\circ$ videos. This fine-tuning task is effort-intensive and requires several transformation techniques to adapt the  distortion~\cite{ktn,tangent}. An intuitive solution is to fine-tune perspective-based deep networks under omnidirectional supervised data. However, this brute-force migration of perspective-based networks often requires enormous supervision and still leads to significant performance degradation~\cite{revisiting}.

We address the first challenge of reliable benchmark dataset shortage by proposing a new dataset named FLOW360. To the best of our knowledge, this is the first perceptually natural-synthetic 360$^\circ$ video dataset collected for omnidirectional flow estimation. Currently, existing omnidirectional datasets face two significant issues i.e., lack of full 360$^\circ$ FOV (field of view) and lack of perceptual realism. Specifically, OmniFlow\cite{omniflow} dataset only has 180$^\circ$ FOV failing to address the omnidirectional nature, while the dataset proposed in OmniFlowNet\cite{omniflownet} lacks perceptual realism in scene and motion. Meanwhile, perspective optical flow datasets such as~\cite{middlebury,sintel,kitti} have facilitated researchers in investigating perspective optical flow estimation methods~\cite{liteflownet,flownet,flownet2,maskflownet,raft}, where the availability of such omnidirectional videos dataset is essential to advance this particular field. It is worth noting that FLOW360 dataset can be used in various other areas such as continuous flow estimation in 3-frame settings with forward and backward consistency~\cite{selflfow,unflow,mirrorflow}, depth~\cite{omnidepth1,omnidepth2} and normal map estimation~\cite{normal}.

The accommodation to the omnidirectional nature generally requires modification of convolution layers and further refinements on the target dataset due to the presence of radial distortions~\cite{revisiting}, which is caused by projecting 360$^\circ$ videos (spherical) to an equirectangular plane. Existing works design various convolution layers to address the distortion problem, such as spherical convolution~\cite{ktn,spherical1,spherical2,spherical3}, spectral convolution~\cite{spectral1,spectral2} and tangent convolution \cite{tangent}. Although these methods can achieve better performance than classical CNNs, they require immense effort with layer-wise architecture design, which is impractical for high-demanding deployment in the real-world setting.

Instead of adding new convolution layers, we design a novel SLOF (\textbf{S}iamese representation \textbf{L}earning for \textbf{O}mnidirectional \textbf{F}low) framework (Fig.~\ref{fig:architecture}), which leverages the rotation-invariant property of omnidirectional videos to address the radial distortion problem. The term rotation-invariant here implies that 360$^\circ$ videos are rotated in a random projection such that the reverse rotation of such projection is equal to the original projection. This rotation-invariant property ensures that omnidirectional videos can be projected to a planar representation with infinite projections by rotating the spherical videos on three different axis $(X,Y,Z)$, namely ``pitch'', ``roll'' and ``yaw'' operations preserving overall information. Specifically, we design a siamese representation learning framework for learning omnidirectional flow from a pair of consecutive frames and their rotated counterparts, assuming that the representations of these two cases are similar enough to generate nearly identical optical flow in the spherical domain. Besides, we design and compare different combinations of rotational augmentation and derive guidelines for selecting the most effective augmentation scheme.

To summarize, we make three major contributions in this paper: (\textbf{i}) we introduce FLOW360, a new optical flow dataset for omnidirectional videos, to fill the dataset's need to advance the omnidirectional flow estimation field. (\textbf{ii}) We propose SLOF, a novel framework for optical flow estimation in omnidirectional videos, to mitigate the cumbersome framework adjustments for omnidirectional flow estimation. (\textbf{iii}) We demonstrate a new distortion-aware error measure for performance analysis that incorporates the relative error measure based on distortion. Finally, we compare our method with existing omnidirectional flow estimation techniques via kernel transformation~\cite{ktn} to address radial distortions. The FLOW360 dataset, the SLOF framework, and our experimental results provide a solid foundation for future exploration in this important field.
	\section{Related Work}
\label{sec:relatedwork}
\noindent \textbf{Optical Flow Datasets.} Perspective datasets such as \cite{barron,middlebury,mccane,real1,real2,spectra1} comprise synthetic image sequences along with synthetic and hand-crafted optical flow. However, these datasets fall short in terms of perceptual realism and complexities. Even though several optical flow datasets have been published recently in \cite{outdoor,kitti,kitti2012,kitti2015}, they are primarily used in automotive driving scenarios. The other relevant dataset in the literature was Sintel~\cite{sintel}, which provided a bridge to contemporary optical flow estimation and synthetic datasets that can be used in real-world situations.

All datasets, as mentioned earlier, are introduced for perspective videos thus cannot be used for omnidirectional flow estimation. So to address this problem, LiteFlowNet360~\cite{revisiting} on omnidirectional flow estimation was released to augment the Sintel dataset by introducing distortion artifacts for the domain adaptation task. Nevertheless, these augmented datasets are discontinuous around the edges and violate the 360$^\circ$ nature of omnidirectional videos. The closest datasets to ours are OmniFlow~\cite{omniflow} and OmniFlowNet~\cite{omniflownet}. OmniFlow introduced a synthetic 180$^\circ$ FOV dataset, which is limited to indoor scenes and lacks full 360$^\circ$ FOV. Similarly, OmniFlowNet introduced a full 360$^\circ$ FOV dataset. However, both datasets lack complexities and evidence for perceptual realism. We show a detailed comparison of FLOW360, OmniFlow, and OmniFlowNet in Fig.~\ref{fig:datastat}. Compared to existing datasets in the literature, FLOW360 is the first perceptually natural benchmark 360$^\circ$ dataset and fills the void in current research.

\noindent \textbf{Optical Flow Estimation.} Advancements in optical flow estimation techniques largely rely on the success of data-driven deep learning frameworks. Flownet~\cite{flownet} marked one of the initial adoption of CNN- based deep learning frameworks for optical flow estimation. Several other works~\cite{flownet2,liteflownet,flow0,flow1,flow2,flow3,flow4,flow5} followed the footsteps with improved results. Generally, these networks adopt an encoder-decoder framework to learn optical flow in a coarse-to-fine manner. The current framework RAFT~\cite{raft} has shown improvements with correlation learning.

The methods mentioned above are insufficient on omnidirectional flow field estimation as they are designed and trained for perspective datasets. One of the initial work~\cite{backproject} on omnidirectional flow estimation was presented as flow estimation by back-projecting image points to the virtually curved retina, thus called back-projection flow. It showed an improvement over classical algorithms. Similarly, another classical approach~\cite{wavelet} relyed on spherical wavelet to compute optical flow on omnidirectional videos. However, these methods are limited to classical approaches as they are not relevant in existing deep learning-based approaches. One of the recent works, LiteFlowNet360~\cite{revisiting} tried to compute optical flow on omnidirectional videos using domain adaptation. This method utilized the kernel transformer technique (KTN~\cite{ktn}) to adapt convolution layers on LiteFlowNet~\cite{liteflownet} and learn correct convolution mapping on spherical data. Similarly, OmniFlowNet~\cite{omniflownet} proposed a deep learning-based optical flow estimation technique for omnidirectional videos. The major drawback of these methods is the requirement to adapt convolution layers, which takes a substantial amount of time and makes portability a significant issue. For example, in LiteFlowNet360, each convolution layer in LiteFlowNet was transformed using KTN with additional training and adjustments. Similar to OmniFlowNet, every convolution layer in LiteFlowNet2~\cite{liteflownet2} was transformed using kernel mapping~\cite{kmapping} based on different locations of the spherical image. These techniques incur computational overheads and limit the use of existing architectures. Such approaches demand explicit adaptation of convolution layers, which is hard to maintain when more up-to-date methods are published constantly. Contrary to these methods, we propose a Siamese Representation Learning for Omnidirectional Flow (SLOF) method to learn omnidirectional flow by exploiting existing architectures with designed representation learning objectives, significantly reducing the unnecessary effort of transforming or redesigning the convolution layer.

\noindent \textbf{Siamese Representation Learning.} Representation learning is a powerful approach in unsupervised learning. Siamese networks have shown great success in different vision-related tasks such as verification~\cite{signature,verification,oneshot} and tracking\cite{tracking}. A recent approach~\cite{simsiam} in siamese representation learning showed impressive results in unsupervised visual representation learning via exploiting different augmentation views of the same data. They presented their work in pre-training and fine-tuning stages, where the former being the unsupervised representation learning. We use the representation learning scheme on omnidirectional data via rotational augmentations, maximizing the similarity for latent representations and minimizing the flow loss.
	\section{FLOW360 Dataset}
\label{sec:dataset} 
\begin{figure*}[!t]
\centering
    \includegraphics[width=\linewidth]{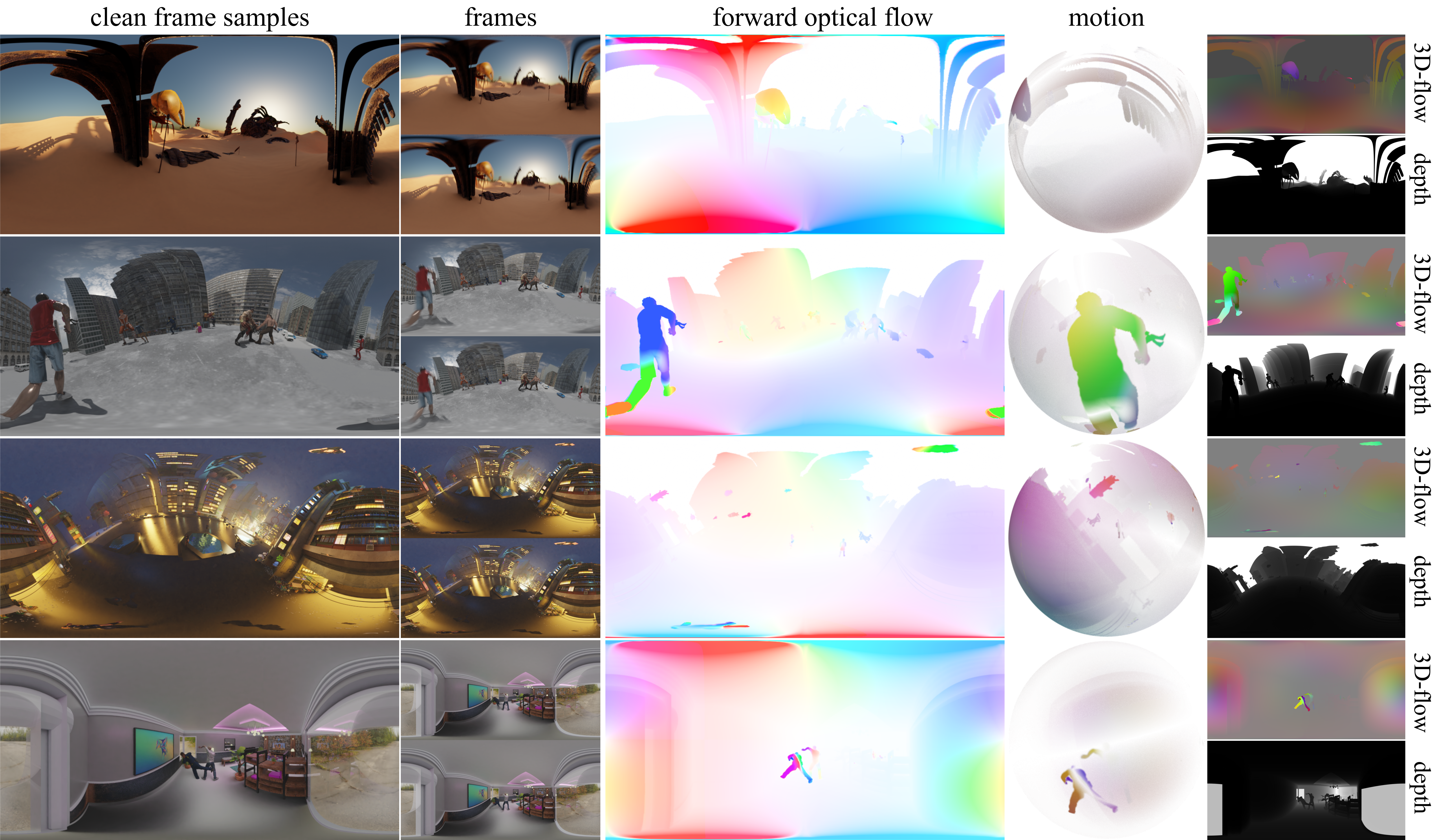}
    \caption{\textbf{The FLOW360 Dataset}. Sample frames (first and second column, respectively) from some of the videos with corresponding forward optical flow and dynamic depth information. Motion in 3D Sphere (fourth column) is computed by transforming the motion vectors from Equirectangular plane $(\theta, \phi)$ to unit sphere $f(x,y,z)$. Motion in the sphere is represented in RGBA color notation. RGB color representation (as suggested in Middlebury~\cite{middlebury}) is encoded using $(x,y)$ components, and the alpha color is encoded from $z$ of a unit sphere. RGB encoding (fifth column) is an RGB color map of flow in 3D space. \textbf{Note}: flow fields are clipped for better visualization.}
    \label{fig:samples}
    \vspace{-0.1in}
\end{figure*}
\begin{figure}[!t]
\centering
    \includegraphics[width=\linewidth]{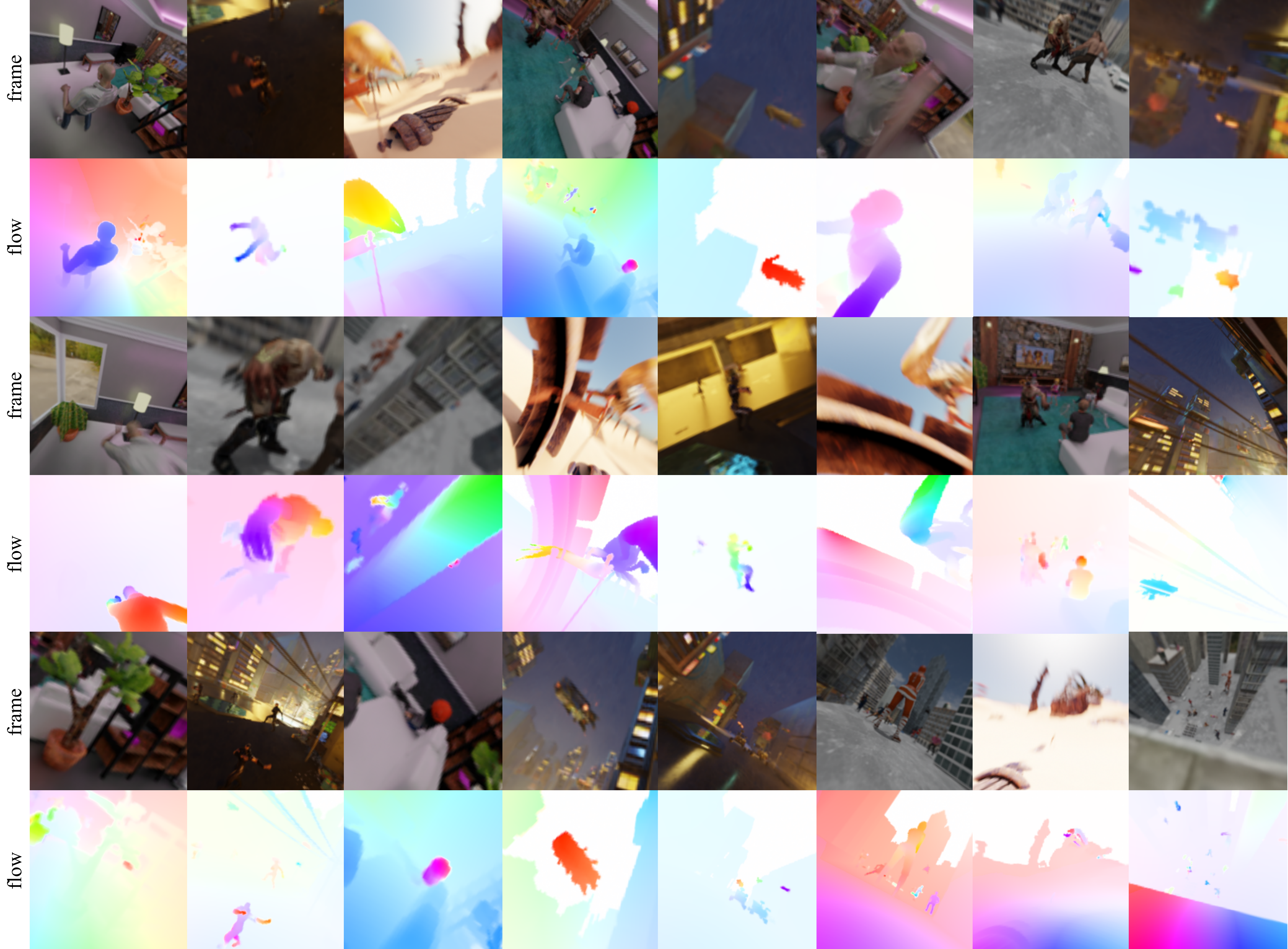}
    \caption{\textbf{Motion and Scene Diversity}. Samples from FLOW360 Dataset with random projection (pitch, roll, yaw, fov) showing scene and motion diversity. The FLOW360 dataset has a vast scene consisting of several lighting scenarios, textures, diverse 3D assets, and motion complexity in different regions.}
    \label{fig:pers_flow}
    \vspace{-0.1in}
\end{figure}
FLOW360 is an optical flow dataset tailored for 360$^\circ$ videos using Blender~\cite{blender}. This dataset contains naturalistic 360$^\circ$ videos, forward and backward optical flow, and dynamic depth information. The dataset comprises 40 different videos extracted from huge 3D-World `The Room’, `Modern’, `Alien Planet’, and `City Rush’. Due to their size, this 3D-World cannot be rendered at once in a single video. We render several parts of this 3D-World, which provides enough qualitative variation in motion and visual perception like 3D-assets, textures, and illuminations. The nature of this large and diverse animated world provides relatively enough diversity to qualify for a standard benchmark dataset. The Fig.~\ref{fig:pers_flow} shows some of the examples of motion and scene diversity of FLOW360. Similarly, samples from the dataset of different 3D-World are shown in Fig.~\ref{fig:samples}. We build these 3D-World using publicly available 3D models~\cite{alienplanet,cityrush,modern} and 3D animated characters~\cite{turbosquid,sketchfab,mixamo}. Meanwhile, we adopt Blender~\cite{blender} for additional rigging and animation for the dataset.

FLOW360 contains 40 video clips extracted from different parts of huge 3D-World, `The Room’, `Alien Planet’, `City Rush’, and Modern’. The datasets also contain other information like depth maps and normal fields extracted from the 3D-World. The FLOW360 dataset has 4,000 video frames, 4,000 depth maps, and 3,960 flow fields. We divide the video frames into 2700/1300 train/test split. We render the video frames with the dimension of $(512,1024)$ to save the rendering time. However, FLOW360 can be rendered with higher resolution, as 3D models and Blender add-ons will also be public.

\begin{figure}[!t]
\centering
    \includegraphics[width=0.99\linewidth]{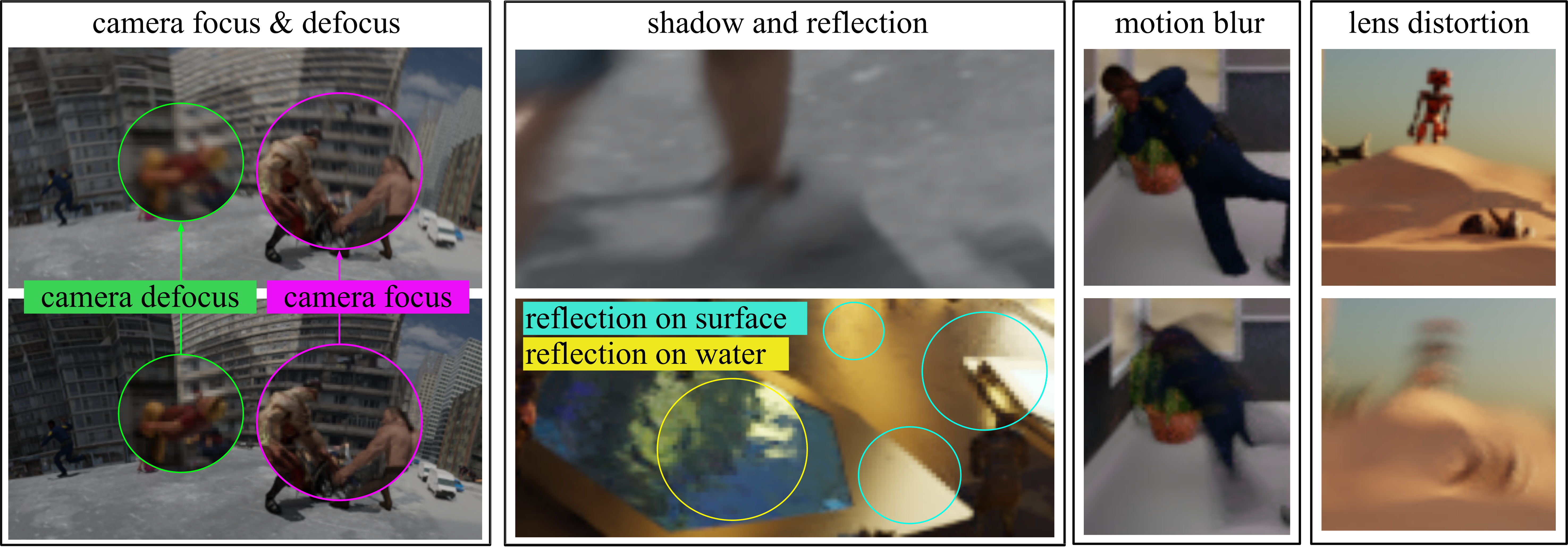}
    \caption{\textbf{Complexity of FLOW360 Dataset}. Final frames in FLOW360 Dataset include complex characteristics like camera focus/defocus, motion blur, lens distortion, shadow, and reflections. Our dataset provides ambiance occlusion and environmental effects for a realistic visual appearance.}
    \label{fig:complexity}
    \vspace{-0.2in}
\end{figure}

\begin{figure*}[!t]
\centering
    \includegraphics[width=\linewidth]{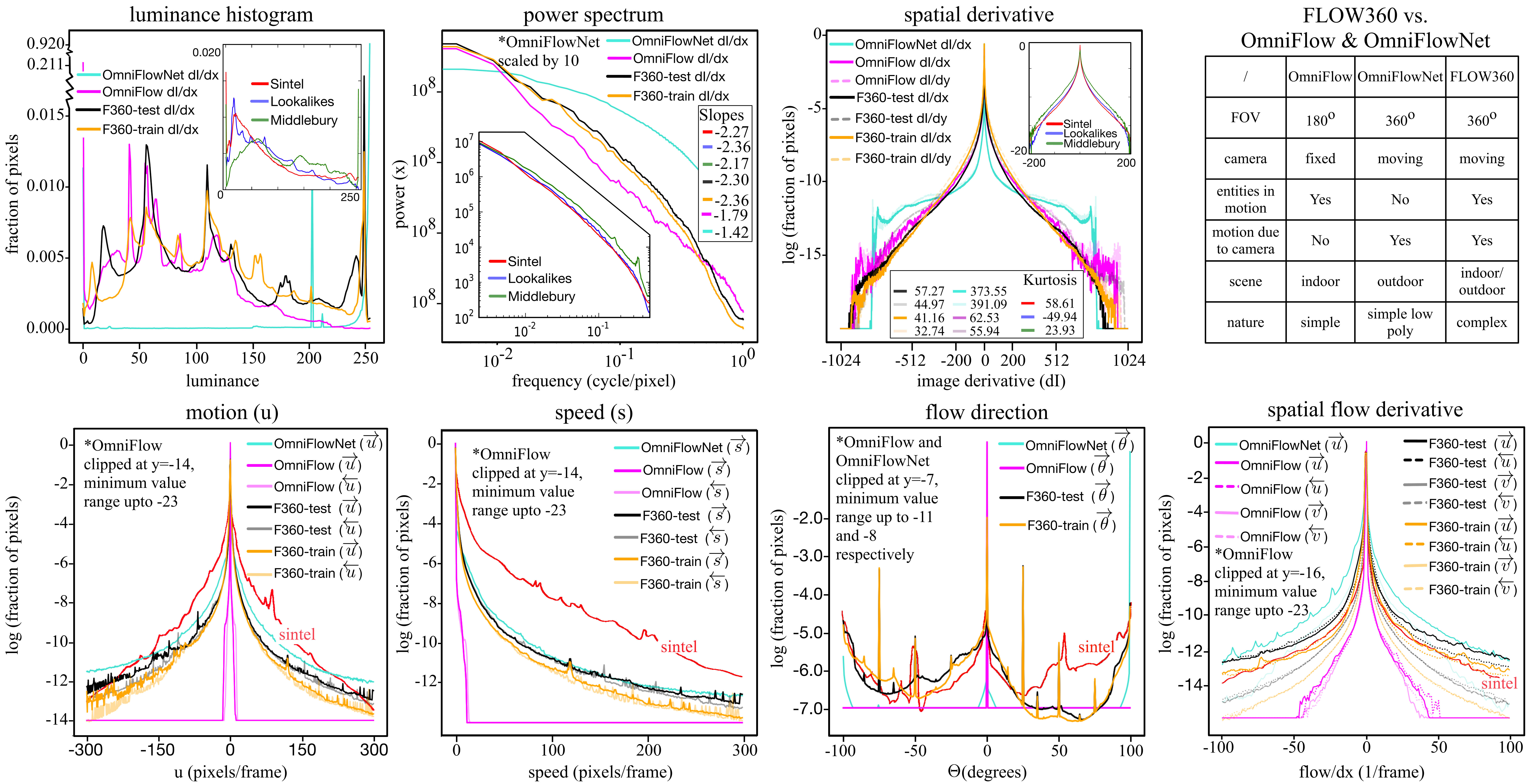}
    \caption{\textbf{Comparision of frames and flow statistics}.  Top row represents the frames statistics and comparison with Sintel, Lookalikes, Middlebury, OmniFlow~\cite{omniflow} and OmniFlowNet~\cite{omniflownet}. Bottom row represents flow statistics and comparison with Sintel (\textcolor{red}{red}), OmniFlow (\textcolor{magenta}{magenta}) and OmniFlowNet (\textcolor{turquoise}{turquoise}). The table on the top-right shows a brief comparision of OmniFlow \& OmniFlowNet with FLOW360 dataset. \textbf{Note}: ($\rightarrow$, $\leftarrow$) represents forward and backward flow fields, respectively.}
    \label{fig:datastat}
    \vspace{-0.2in}
\end{figure*}

\noindent \textbf{Diversity.} We design FLOW360 datasets to include a diverse situation that resembles the real world scenario as much as possible. The statistical validity of the datasets in terms of perceptual realism of scene and motion is presented in Fig.~\ref{fig:datastat}. The datasets contain a wide range of motion complexity from smaller to larger displacement, occlusion, motion blur, and similar complexities on the scene using camera focus-defocus, shadow, reflections, and several distortion combinations. As these complexities are quite common in natural videos, the FLOW360 provides similar complexities. Similarly, the datasets cover diverse scenarios like environmental effects, textures, 3D assets, and diverse illuminations. The qualitative presentation of these diversities and complexities are presented in Fig.~\ref{fig:pers_flow} and Fig.~\ref{fig:complexity} respectively. 

\noindent \textbf{Fairness.} The FLOW360 dataset contains custom-tailored animated 360 videos. We plan to release the dataset with the 3D models and our custom Blender add-ons to provide researchers a platform to create their custom optical flow datasets for all kinds of environments (perspective, 180$^\circ$ and 360$^\circ$ FOV). However, the release of 3D world scenes can raise questions regarding fairness. To mitigate this issue, we will perturb certain parts of 3D world scenes and not release any camera information related to the test set.

\noindent \textbf{Flow-generator with Blender Add-ons.} Flow-generator is a custom Blender add-on written for Blender-v2.92. The flow-generator serves two basic purposes. First, it creates a Blender compositor pipeline to collect frames, depth maps and optical flow information. This add-on can also collect additional information, such as normal maps. Second, it sets up a camera configuration for 360$^\circ$ FOV. We will describe details of the add-ons in supplementary material.

\noindent\textbf{Render Passes.} We exploit several modern features from Blender-v2.92 like advanced ray-tracing as a render engine along with render passes like vector, normal, depth, mist, and so on to produce realistic 3D scenes. Additionally, we incorporate features like ambient occlusion, motion blur, camera focus/defocus, smooth shading, specular reflection, shadow, and camera distortion to introduce naturalistic complexity (shown in Fig.~\ref{fig:complexity}) in our dataset. Besides optical flow information, the FLOW360 3D-world may be used to collect several other helpful information like depth, normal maps, and semantic segmentation.

\noindent \textbf{Dataset Statistics.} We conduct a comprehensive analysis and compare our dataset with Sintel~\cite{sintel}, Lookalikes (presented in the original Sintel paper to compare the image statistics with the simulated dataset), Middlebury~\cite{middlebury}, OmniFlow~\cite{omniflow} and OmniFlowNet~\cite{omniflownet}. The analysis shown in Fig.~\ref{fig:datastat} shows the image and motion statistics in the top and bottom rows, respectively.

Based on analysis from Sintel, we present frame statistics with three different analysis: luminance histogram, power spectrum, and spatial derivative. For luminance statistics, we convert the frames to gray-scale, $I(x,y){\in}[0,255]$ then we compute histograms of gray-scale images across all pixels in the entire dataset. The luminance statistics show the FLOW360 has a similar distribution with the peak in the range between $[0{-}100]$ and decreasing luminosity beyond that range. Similarly, we estimate power spectra from the 2D FFT of the $512{\times}512$ in the center of each frame. We compute the average of these power spectra across all the datasets. We present power spectra analysis separately for the training and test set in this analysis. The power spectra analysis closely resembles the Sintel, Lookalikes, and Middlebury datasets. Based on~\cite{spectra1,spectra2}, the real-world movies exhibit a characteristic of a power spectrum slope around -2, which is equivalent to a $1/f^{2}$ falloff. FLOW360 with the slope $(-2.30, -2.36)$ on test and training split shows such characteristics. We do not claim that FLOW360 is realistic, but it certainly exhibits perceptual similarity with natural movies. The spatial and temporal derivative analysis additionally supports this characteristic. The Kurtosis of frames spatial derivatives range from 32.74 to 57.27, peaked at zero. This characteristic shows that FLOW360 has a resemblance to natural scenes~\cite{spectra1}.

Regarding the flow field analysis we directly compare the distribution of motion $u(x,y)$, speed defined as $s(x,y){=}\sqrt{u(x,y)^2 + v(x,y)^2}$, flow direction $\Theta(x,y){=}\tan^{-1}{(v(x,y)/u(x,y))}$ and spatial flow derivative of $u$ and $v$. The close resemblance of the flow field statistics between Sintel and FLOW360 suggests motion field resemblance with natural movies. Based on these comparisons, FLOW360 exhibits sufficient properties evident enough for its perceptual realism and complexities.

\noindent \textbf{Comparison with OmniFlow and OmniFlowNet.} OmniFlow~\cite{omniflow} presents an omnidirectional flow dataset that is roughly similar to FLOW360. However, the major distinction between these datasets is the FOV. FLOW360 provides immersive 360$^\circ$ FOV, whereas OmniFlow provides only 180$^\circ$ FOV showing FLOW360 compared to OmniFlow is the true omnidirectional dataset. Similarly, OmniFlowNet~\cite{omniflownet} presents synthetic omnidirectional flow dataset with 360$^\circ$ FOV. However, this dataset contains low poly unnatural scenes, which can be explained by relatively larger kurtosis $(373.55,391.09)$, characteristic of a power spectrum and luminance distribution (peaked at 255). The overall statistical analysis reveals FLOW360’s better perceptual realism and diversity.

\noindent \textbf{Applications.} As we mentioned, the FLOW360 dataset contains frames and forward flow field and includes backward flow field, depth maps, and 3D-FLOW360 worlds, providing potential for applications like continuous flow-field estimation in 3 frames setting. Besides optical flow estimation, the FLOW360 dataset can be used in other applications such as depth and normal field estimation. Moreover, given 3D-FLOW360 animation data, the researcher can create as many optical flow datasets as needed.
	\section{SLOF}
\label{sec:method}
SLOF, as shown in Fig.~\ref{fig:architecture}, is inspired by the recent work on Siamese representation learning~\cite{simsiam}. Since the method we rely on acts as a hub between several methods like contrastive learning, clustering, and siamese networks, it exhibits two special properties required for our case. First, this method has non-collapsing behavior. Here, the term collapsing refers to a situation where an optimizer finds possible minimum -1 similarity loss resulting degenerate solution (characterized by zero $std$ of $l_2$-normalized output $z/||z||_2$ for each channel) while training without stop-gradient operations. Stop-gradient yields $std$ value near $\frac{1}{\sqrt{d}}$ across each channel for all samples preventing such behaviour~\cite{simsiam}. Second, it is useful when we have only positive discriminative cases. SLOF does not consider radial distortion mitigation via changing/transforming the convolution layers rather learns the equivariant properties of 360 videos via siamese representation. We claim that such transformation is trivial, based on the following fact. First, the omnidirectional videos are projected in angular domain, \textit{w.r.t.} $\textbf{polar}(\theta),\, \textbf{azimuthal}(\phi); \theta{\in}(-\frac{\pi}{2}, \frac{\pi}{2}), \phi{\in}(-\pi, \pi)$, so we can learn flow fields in these domains and convert these flow fields to spherical domain using planar to spherical transformations as shown in Eq.~\ref{eq:planetosphere} and Eq.~\ref{eq:spheretoplane}. Second, the intent of a convolution operator in optical flow architecture is relatively different from other applications like classification, detection, or segmentation network, where other tasks require convolution to learn relevant features (spatially consistent), the relevance of these features should stay consistent (strictly for better performance) throughout any spatial location of the images/videos. However, the convolution operation is dedicated to computing the pixel-wise displacement regardless of spatial inconsistency in the distorted region via equivariant representation learning~\cite{simsiam}. Another important consideration of such a design is to make this method portable to any existing optical flow architecture. This design will eliminate the cumbersome architecture re-adjustments tasks and make it powerful and portable.

\noindent \textbf{Mapping Flow Field to Unit Sphere.} Input to our model are equirectangular images projected in angular domain $\textbf{polar}(\theta),\textbf{azimuthal}(\phi)$, where these angles are defined in radian as $\theta{\in}(-\frac{\pi}{2}, \frac{\pi}{2}),\phi{\in}{(-\pi, \pi)}$, thus the predicted optical flow is in $(\theta, \phi)$. These flow fields can be converted to unit sphere using planar to spherical co-ordinate transformation as shown below:
\begin{equation}
  (x_s,y_s,z_s) = (\sin{\theta}\cos{\phi}, \sin{\theta}\sin{\phi},\cos{\theta}).
  \label{eq:planetosphere}
\end{equation}
We can compute sphere to catadioptric plane~\cite{catdiop} projections to express the flow field in Cartesian co-ordinates as:
\begin{equation}
 (x,y) = (\frac{x_s}{1-z_s},\frac{y_s}{1-z_s}) = (\cot{\frac{\theta}{2}\cos{\phi}}, \cot{\frac{\theta}{2}\sin{\phi}}).
  \label{eq:spheretoplane}
\end{equation}

\noindent\textbf{Design.}~Given a pair of input image sequence $X_1{=}(x_1,x_2)$, the rotation head $(R)$ computes augmented view of this sequence as $X_2{=}(x^{\prime}_1, x^{\prime}_2)$ with rotation $r$ using a random combination of ``pitch", ``yaw" and ``roll" operations. These two augmented views are passed as an input to an encoder network $f$, defined as $f{=}{P(R^\prime(\Theta(E(R(X, r)))))}$ where $E$ is a flow prediction module, RAFT~\cite{raft} in our case, $\Theta$ is a mapping of 2D flow to unit sphere, $R^\prime$ is a reverse rotation operation and $P$ is a convolution based down-sampling head. A prediction head presented as $h$ (an MLP head), transforms the output from the encoder $f$ from one stream to match the other stream. The illustration of this process shown in Equation.~\ref{eq:cosine} as maximization of cosine similarity two views from siamese stream:
\begin{equation}
  D(p^{left}, z^{right}) =- \frac{p^{left}}{||p^{left}||_2} \cdot \frac{z^{right}}{||z^{right}||_2}.
  \label{eq:cosine}
\end{equation}
Here, $p^{left}{\triangleq}h(f^{left}(X_1))$ and $z^{right}{\triangleq}f^{right}(X_2)$ denotes the output vectors to match from two different streams$(f^{left},f^{right})$. This maximization problem can be viewed from another direction, with $(p^{right}, z^{left})$ as the second matching pair from siamese stream $(f^{right},f^{left})$ respectively. Given two matching pairs, we can use following (Eq. \ref{eq:simloss}) symmetrized similarity loss function (note that $z^{left}$ and $z^{right}$ are treated as a constant term using stop-grad operations to prevent a degenerate solution due to model collapse~\cite{simsiam}):
\vspace{-0.05in}
\begin{equation}
  \vspace{-0.05in}
  L_{sim} =\frac{1}{2}D(p^{left},z^{right})+\frac{1}{2}D(p^{right},z^{left}).
  \label{eq:simloss}
  \vspace{-0.05in}
\end{equation}
Similarly, the optical flow loss is computed as a sequence loss~\cite{raft} over predicted flow field and ground truth. This loss ($l_1$ distance over predicted and ground truth flow $f_{gt}$) is computed and averaged over sequence of predictions iteraterively generated for the same pair of input frames $\{f_1, f_2, ..., f_n\}{=}E(R(X,r))$ as shown in Eq.~\ref{eq:epeloss}, where $\gamma{=}0.8^{n - i - 1}$ served as weights over sequence loss. Note that $(n,i)$ denotes number of prediction$(n)$ in sequence and prediction id$(i)$ in predicted flow sequences. The design of the weighted schemes ensures different levels of confidence on predicted flows over time.
\vspace{-0.1in}
\begin{equation}
  \vspace{-0.05in}
  L_{flow} =\sum_{i=1}^{n}{\gamma||R(f_{gt},r) - f_i||}.
  \label{eq:epeloss}
  \vspace{-0.05in}
\end{equation}
Given similarity loss$(L_{sim})$ and flow loss$(L_{flow})$ we implement a hybrid loss function $L{=}L_{sim}{+}L_{flow}$. The overall objective of this loss function is to maximize the similarity between latent representation of flow information while minimizing the loss between ground truth and predicted optical flow.
\begin{figure*}[!t]
    \centering
    \includegraphics[width=\linewidth]{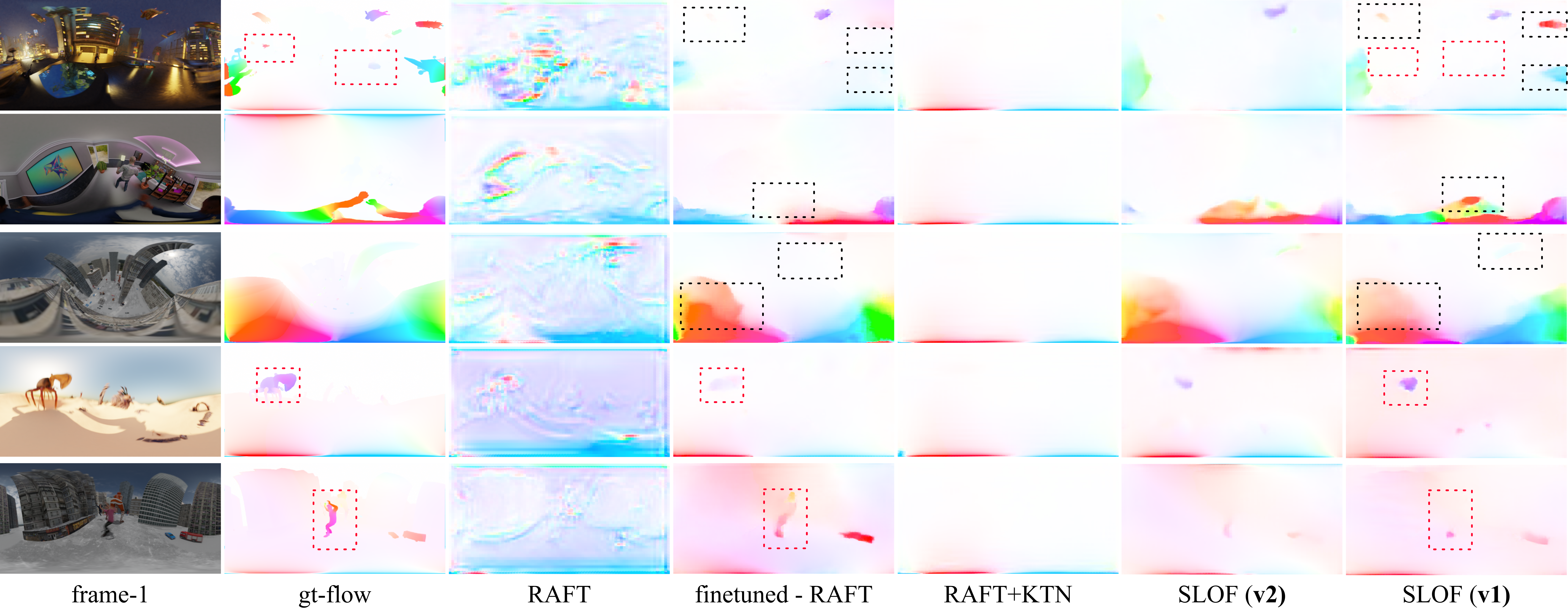}
    \caption{\textbf{Qualitative results on FLOW360 test set}.  Qualitative results show our best model SLOF(\textbf{v1}) shows better results compared to fine-tuned RAFT trained with policy explained in~\cite{raft}. The dotted (\textbf{black}) rectangle indicates the comparative improvements of our model over fine-tuned RAFT. RAFT+KTN method fails to predict flow-field correctly; instead, it only predicts shallow flow fields from camera motion. The weakness of our model can be seen on dotted (\textcolor{red}{red}) rectangle where smaller motion segments are missing. \textbf{Note:} Flows information is clipped for better visualization.}
    \label{fig:qualitative}
    \vspace{-0.2in}
\end{figure*}

	\section{Experiments}
\label{sec:experiment}
We evaluate SLOF on the FLOW360 test set. We use pre-trained RAFT on Sintel~\cite{sintel} and fine-tune on FLOW360 as a comparison baseline. The fine-tuning process is done using training protocols suggested in \cite{raft}. Moreover, to make a fair comparison with traditional methods, we transform RAFT (pre-trained) to adapt spherical convolution using KTN~\cite{ktn}. KTN transforms the convolution kernel to mitigate the radial distortions via estimating the spherical convolution function. Additionally, we run ablation studies on different training strategies and propose a distortion-aware evaluation. We will present details of the training procedure in the supplemental material.
\begin{figure}[!t]
\centering
    \includegraphics[width=0.9\linewidth]{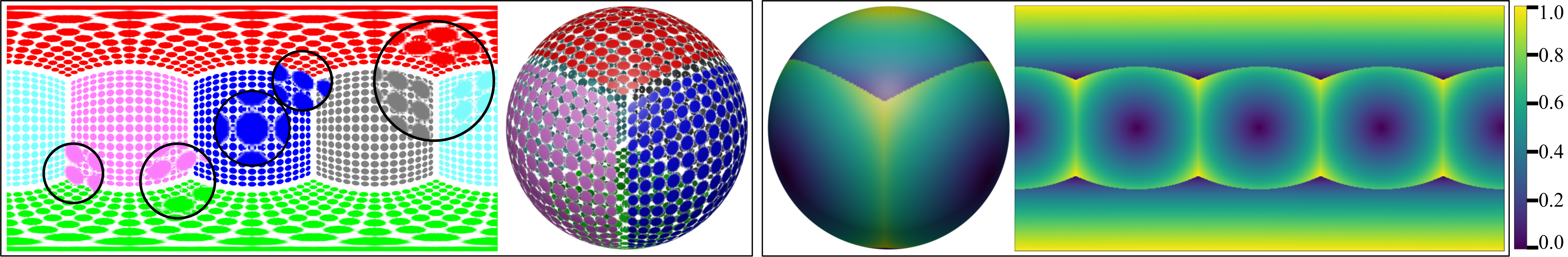}
    \caption{\textbf{Distortion density map}.  Illustrating different distortion intensity due to equirectangular projections. Left: upper (\textcolor{red}{red}) and lower (\textcolor{green}{green}) part of projections shows higher distortion in central part where as the equatorial region (\textcolor{cyan1}{cyan}, \textcolor{pink1}{pink}, \textcolor{blue}{blue}, \textcolor{gray}{gray}) exhibit higher distortion rate away from the center of tangential plane. Right: shows the distortion density from $(0,1)$. This distortion density map is used to evaluate the distortion aware EPE (EPE$_d$). \textbf{Note}: Each circle patch in left spherical projection have same area.}
    \label{fig:distortiondensity}
    \vspace{-0.2in}
\end{figure}

\noindent\textbf{Scope.} The scope of our experiments are two folds: First, create a baseline for future researchers to explore novel methodologies. Second, address the validity of our method based on the fair comparison with a flow network designed for a spherical dataset. We formulate our baseline experiment on perspective optical flow network RAFT and modified version of RAFT with KTN~\cite{ktn} to compare the performance. The RAFT+KTN architecture simulates a domain adaptation similar to approaches like \cite{revisiting,omniflownet}. We choose KTN because of its success over alternative approaches like~\cite{spectral1,spectral2,saliency,spherical2,spherical3}. It is worth noting that the design of omnidirectional flow estimation can be extended to several techniques involving mitigation of radial distortions, making it practically impossible to cover all.

\noindent\textbf{Augmentation Strategy.} Given the nature of SLOF, we can train it using two different training strategies (\textbf{v1},\textbf{v2}) as shown in Fig.~\ref{fig:architecture}(right). These strategies can be achieved by performing different rotational augmentation on the input sequences. The first strategy (\textbf{v1}) can be achieved by using set of inputs $(R(X_1,r_1), R(X_2,r_2))$ where $r_1{=}(0,0,0)$, i.e., $X_1$ does not have any rotational augmentation, whereas $r_2{\neq}(0,0,0)$ has rotation defined with random combinations of ``pitch", ``roll", and ``yaw" operations. This setting is kept consistent throughout the training process. Alternatively, identical augmentation can be achieved by flipping this augmentation protocols. The second rotational scheme (\textbf{v2}) can be achieved by randomly switching rotation such that when $r_1$ is none, the $r_2$ is some random rotational augmentation and vice versa. This approach performs on par with \textbf{v1}.

\setlength{\belowdisplayskip}{1pt} \setlength{\belowdisplayshortskip}{1pt}
\setlength{\abovedisplayskip}{1pt} \setlength{\abovedisplayshortskip}{1pt}
\begin{equation}
  \text{AE} = \arccos(\frac{u_eu_r + v_ev_r + 1}{\sqrt{u_r^2+v_r^2+1} \sqrt{u_e^2+v_e^2+1}}).
  \label{eq:ae}
\end{equation}
\begin{equation}
  \text{EPE} = \frac{1}{N}\sum_{i}^{N}{{||f_{pred} - f_{gt}||_2}}.
  \label{eq:epe}
\end{equation}
\begin{equation}
  \text{EPE}_{d} = \frac{1}{N}\sum_{i}^{N}{\frac{||f_{pred} - f_{gt}||_2}{1-d}}.
  \label{eq:distortionepe}
\end{equation}
\noindent\textbf{Evaluation Strategy.} We evaluate our method based on 2D-raw flow. Besides, using EPE (End Point Error in Eq.~\ref{eq:epe}), i.e., Euclidean distance between the predicted flow and ground truth flow, as a single evaluation metric, we incorporate AE (Angular Error) as shown in Eq.~\ref{eq:ae} as the second measure. To explain the error in the omnidirectional setting, we introduce a distortion-aware measure called EPE$_d$ as in Eq.~\ref{eq:distortionepe}. This metric penalizes the error in the distorted area based on the distortion density map.

As EPE$_d$, AE$_d$ is calculated as $\frac{1}{N}\sum_{i}^{N}{\frac{\text{AE}}{1-d}}$ where, $d$ represents the distortion density map illustrated in ~Fig.~\ref{fig:distortiondensity}, $f_{pred}{=}(u_e,v_e)$ represents predicted flow, and $f_{gt}{=}(u_r,v_r)$ represents ground truth flow. Note that, to maintain lower metrics scale the distortion density is mapped between $[0.500,1.000)$  from $(0.0, 1.0]$. Please refer to supplemental for additional details on distortion density map.

\begin{figure}[!t]
\centering
    \includegraphics[width=0.99\linewidth]{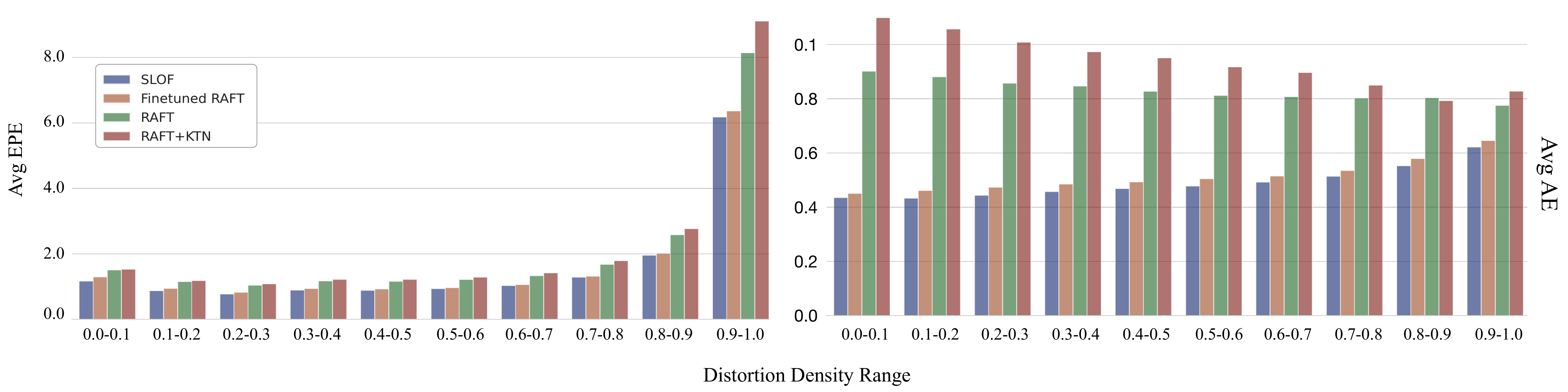}
    \vspace{-0.2in}
    \caption{\textbf{Error distribution plot}. Illustrating error (EPE and AE) in different distortion density ranges. SLOF relatively performs better in all distortion density ranges.}
    \label{fig:histoplot}
    \vspace{-0.2in}
\end{figure}

\begin{table*}[!t]\scriptsize
\caption{\textbf{Quantitave results on FLOW360 test set.} $\ast$ denotes that we use EPE$_d$/AE$_d$ as the metrics; otherwise, the normal EPE and AE. Compared to baseline, SLOF achieves lower end-point-error and angular error on both distortion aware (EPE$_d$ and AE$_d$) and normal scheme. In terms of end-point-error (lower the better) our model (\textbf{v1},\textbf{v2}) outperforms all the baseline. Similarly in terms of angular error (lower the better) our models (\textbf{v1}, \textbf{v2}) perform comparatively similar and outperform all the baseline. Though RAFT+KTN achieves comparable normal EPE, the distortion aware (Weighted) metrics (EPE$_d$ and AE$_d$) are significantly larger. \textbf{Note}: metrics in range (all, less than (5, 10, 20) and greater than 20) is computed as an average, based on the speed (\textbf{s}(x,y)${=}\sqrt{u(x,y)^2+v(x,y)^2}$) only in the respective pixel regions.}

\centering
\begin{tabular}{c|c|ccccccccc}
\toprule
\rowcolor{white}
Mehtod & Version & Metric & Weighted \textbf{s}$\geq$0$^\ast$ & \textbf{s}$\geq$0 & \textbf{s}$<$5 & \textbf{s}$<$10 & \textbf{s}$<$20 & \textbf{s}$\geq$20 \\
\hline
\multirow{6}{*}{Baselines} & \multirow{2}{*}{RAFT~\cite{raft}} & EPE & 3.344 & 2.058 & 0.558 & 0.682 & 0.838 & 71.736 \\
& & \cellcolor{Gray}AE & \cellcolor{Gray}1.120 & \cellcolor{Gray}0.820 & \cellcolor{Gray}0.825 & \cellcolor{Gray}0.821 & \cellcolor{Gray}0.819 & \cellcolor{Gray}0.868  \\
\hhline{~|-|---------}
& \multirow{2}{*}{Finetuned RAFT~\cite{raft}} & EPE & 2.635 & 1.624 & 0.314 & 0.393 & 0.509 & 65.340\\
& & \cellcolor{Gray}AE & \cellcolor{Gray}0.745 & \cellcolor{Gray}0.522 & \cellcolor{Gray}0.527 & \cellcolor{Gray}0.522 & \cellcolor{Gray}0.520 & \cellcolor{Gray}0.647  \\
\hhline{~|-|---------}
& \multirow{2}{*}{RAFT + KTN~\cite{ktn}} & EPE & 3.899 & 2.222 & 0.598 & 0.742 & 0.924 & 76.426\\
& & \cellcolor{Gray}AE & \cellcolor{Gray}2.020 & \cellcolor{Gray}0.912 & \cellcolor{Gray}0.912 & \cellcolor{Gray}0.910 & \cellcolor{Gray}0.911 &  \cellcolor{Gray}1.0114  \\
\hline
\multirow{4}{*}{SLOF} 
& \multirow{2}{*}{Switch rotation (\textbf{v2})} & EPE & 2.626 & 1.615 & 0.326 & 0.401 & 0.512 & 64.678\\
& & \cellcolor{Gray}AE & \cellcolor{Gray}\textbf{0.691} & \cellcolor{Gray}\textbf{0.485} & \cellcolor{Gray}\textbf{0.489} & \cellcolor{Gray}\textbf{0.484} & \cellcolor{Gray}\textbf{0.482} & \cellcolor{Gray}0.659  \\
\hhline{~|-|---------}
& \multirow{2}{*}{Single rotation (\textbf{v1})} & EPE & \textbf{2.548} & \textbf{1.568} & \textbf{0.309} & \textbf{0.387} & \textbf{0.502} & 62.476\\
& & \cellcolor{Gray}AE & \cellcolor{Gray}0.708 & \cellcolor{Gray}0.497 & \cellcolor{Gray}0.501 & \cellcolor{Gray}0.497 & \cellcolor{Gray}0.495 & \cellcolor{Gray}\textbf{0.607}  \\
\bottomrule
\end{tabular}
  \label{tab:quantitative}
\vspace{-0.2in}  
\end{table*}
\noindent\textbf{Results.} Fig.~\ref{fig:qualitative}, Fig.~\ref{fig:histoplot} and Table~\ref{tab:quantitative} summarize our experimental results. The overall summary of qualitative results is presented in Fig.~\ref{fig:qualitative}. SLOF performs better than baseline RAFT and kernel transformed RAFT+KTN methods. This result is evident enough to show that siamese representation learning can exploit the rotational properties of 360$^\circ$ videos to learn omnidirectional optical flow regardless of explicit architecture adjustments. 

Our methods, SLOF (\textbf{v1},\textbf{v2}) perform better than presented baselines. Among these methods \textbf{v1} has the best EPE score whereas, \textbf{v2} has better AE score. However, AE on both \textbf{v1} and \textbf{v2} are relatively similar, suggesting \textbf{v1} as our best method. This is clearly visible in qualitative results shown in Fig.~\ref{fig:qualitative}. 

By investigating distortion-aware EPE, we can see that RAFT with KTN achieves significantly higher EPE regardless of comparable normal EPE with the other methods. This clearly explains why RAFT+KTN methods could not predict the motion around the distorted area; instead, it predicts shallow flow fields due to camera motion only. Moreover, comparing qualitative results in Fig.~\ref{fig:qualitative} and EPE measure in different distortion ranges in Fig.~\ref{fig:histoplot}, we can see that our best method can predict smoother flow fields compared to baseline methods. These fields in the polar region are comparatively better and have better motion consistency in the edge region. However, our model might fail to predict relatively smaller motion regions in some cases, which leaves room for future improvements based on the proposed method. This concludes that RAFT+KTN requires additional re-engineering and domain adaptation, which is out of the scope of current work.

	\section{Conclusion}
\label{sec:conclusion}
Omnidirectional flow estimation remains in its infancy because of the shortage of reliable benchmark datasets and tedious tasks dealing with inescapable radial distortions. This paper proposes the first perceptually natural-synthetic benchmark dataset, FLOW360, to close the gap, where comprehensive analysis shows excellent advantages over other datasets. Our dataset can be extended for other non-motion applications like segmentation and normal estimation task as well. Moreover, we introduce a siamese representation learning approach for omnidirectional flow (SLOF) instead of redesigning the convolution layer to adapt omnidirectional nature. Our method leverages the invariant rotation property of 360$^\circ$ videos to learn similar flow representation on various video augmentations. Meanwhile, we study the effect of different rotations on the final flow estimation, which provides a guideline for future work. Overall, the elimination of network redesigns aids researchers in exploiting existing architectures without significant modification leading faster deployment in real world setting. \\

	\noindent \textbf{Acknowledgements.} This research was partially supported by NSF CNS-1908658, NeTS-2109982 and the gift donation from Cisco. This article solely reflects the opinions and conclusions of its authors and not the funding agents.
	
	\bibliographystyle{splncs04}
	\bibliography{egbib}
	\clearpage
	
	\title{SLOF Supplemental Material}
\author{}
\authorrunning{K. Bhandari et al.}
%
\institute{}

\maketitle

\section{Flow Generator}

FLOW360 is created using Blender\footnote{\url{https://www.blender.org/}}, free and open source 3D creation suite. Blender provides an interface to write add-ons for automating workflows. We create Flow-generator, an add-on written for Blender-v2.92 to collect optical flow and several other data like depth information and normal maps. Flow-generator can be installed using following steps:
\begin{enumerate}
    \item Download Flow-generator\footnote{\url{https://www.dropbox.com/s/v2mvjvs7ze8rzj1/flowgenerator.tar.gz?dl=0}} 
    \item In Blender-v2.92, follow:
    \begin{itemize}
        \item Edit $\rightarrow$ Preferences $\rightarrow$ Add-ons $\rightarrow$ Install
        \item Select flow\_generator.py from Flow-generator project folder
        \item Install Add-on
        \item Search FlowGenerator
        \item Check mark for installation
    \end{itemize}
    
    \item Setup compositor pipeline in Blender-v2.92
    \begin{itemize}
        \item Go to Compositing
        \item Select ``SetFlow Generator" from Custom Node Group
        \item If Custom Node Group is disabled press ``n" to make it visible
        \item Select desired configuration and click ``SetEnv"
    \end{itemize}
    
    \item Camera Setup
    \begin{itemize}
        \item Go to Layout
        \item Open tab on right side of the Layout view by pressing ``n" if not visible
        \item Go to ``CameraSetup" on the right tab
        \item Select desired configuration and click ``Set Camera System"
    \end{itemize}
    
\end{enumerate}
Note that, we use cloud rendering to speed up the rendering process which requires crowd-render\footnote{\url{https://www.crowd-render.com/}} add-on. Flow-generator provides two basic functionality: (i) camera setup and (ii) compositor pipeline.

\subsection{Compositor Pipeline}
Compositor in Blender provides a pipeline to process render output. Flow-generator creates a basic pipeline (shown in~Fig. \ref{fig:flowgenerator}) to collect information like optical flow, images, depth maps and so on. It also provides an easy way to cache the desired outputs in structured format. Similarly, it also setups basic configurations like selecting render-engine and render passes. 
\subsection{Camera Setup}
\begin{figure*}[!t]
\centering
    \includegraphics[width=\linewidth]{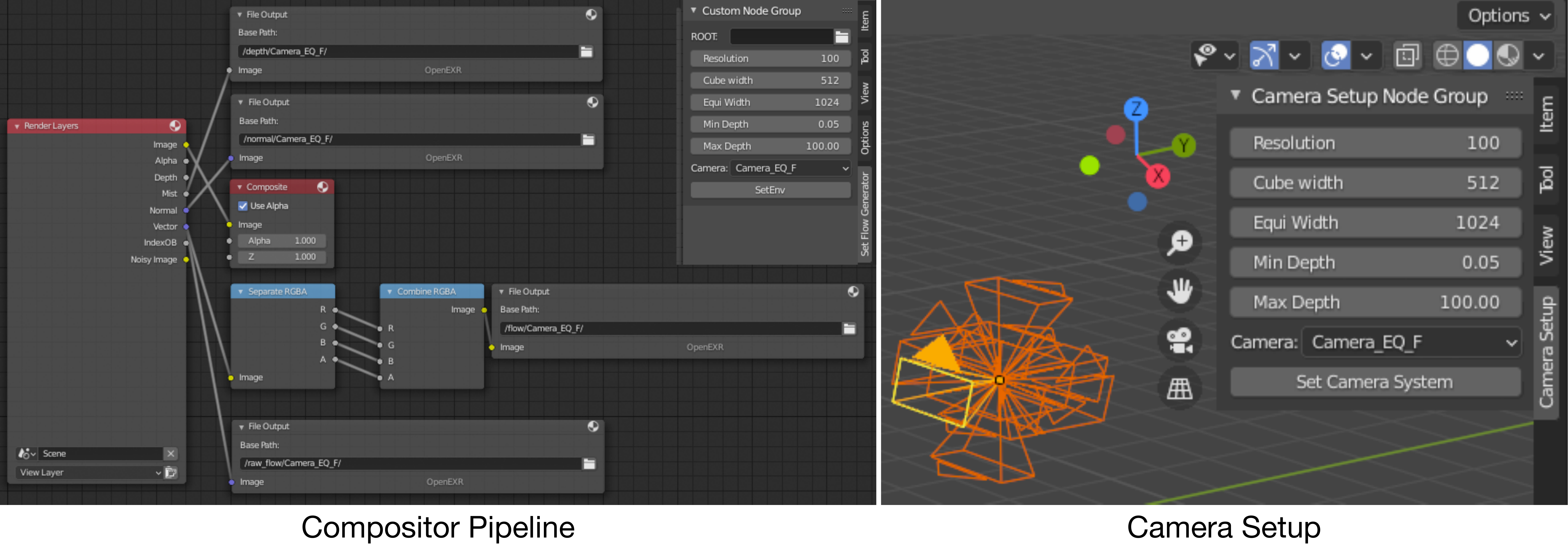}
    \caption{\textbf{FlowGenerator.} Flow-generator serves two major purposes: (Left) Setting up a compositor pipeline which invovles setting up filepaths, pipelining desired output and setting initial configurations like render-engine and render passes. (Right) Setting up camera configurations and additional information like resolution, dimensions and depth of the scene.}
    \label{fig:flowgenerator}
\end{figure*}

Camera Setup (shown in~Fig. \ref{fig:flowgenerator}) can be accessed via 3D-Layout tab in Blender-v2.92 as suggested above. Camera Setup creates an omnidirectional camera with full 360$^\circ$ field of view. The camera setup consists of twelve different cameras (six perspective and six 360$^\circ$) out of which any omnidirectional camera (default Camera\_EQ\_F) can be selected as main camera for rendering omnidirectional videos. It also provides an interface to adjust configurations like resolution, dimension and depth of the scene.

\section{Distortion Density Map}
\label{sec:distortion}
We compute distortion mask (U$_d$,D$_d$,F$_d$,B$_d$,R$_d$,L$_d$) in a cube-map with six faces: Up(U), Down(D), Front(F), Back(B), Right(R) and Left(L) respectively. This distortion density cube-map projection is then projected to equirectangular projection using spherical co-ordinate transformation (shown in Eq.~\ref{eq:planetosphere}).
\begin{equation}
  (x_s,y_s,z_s) = (\sin{\theta}\cos{\phi}, \sin{\theta}\sin{\phi},\cos{\theta}).
  \label{eq:planetosphere}
\end{equation}

To compute density mask (C$_d$, where C$\in$(U,D,F,B,R,L)) in each face, we define a meshgrid for co-oridnates $x$ and $y$ ranging from $([-1,1])$ with dimension of $(256,256)$. The co-ordinates $(x,y)$ are used to compute a radius map (r) of size $(256,256)$ as shown in Eq.~\ref{eq:radius}.

\begin{equation}
  r = \sqrt{x^2 + y^2}
  \label{eq:radius}
\end{equation}

In our paper we have shown that the radial distortion is higher towards central in the polar region which corresponds to (U,D) faces of cube-map projections. Similarly, in equatorial region i.e., rest of the faces (F,R,B,L) shows higher distortion rate away from the center. We compute two distinct distortion map (C$_d$), one for polar regions (U,D) and another for equatorial regions (F,B,R,L) as shown in Eq.~\ref{eq:compute}

\begin{equation}
\label{eq:compute}
C_d =
\begin{cases} 
      1 - r/max(r) & \text{if, C $\in$ \{U,D\} }, \\
      r/max(r) & \text{otherwise}
   \end{cases}
\end{equation}

Please refer to code\footnote{\url{https://www.dropbox.com/s/q1d4eoqvj2a30ij/distortion_weight.ipynb?dl=0}} for additional details.
\section{Experimental Setting}
\label{sec:experiment}
We conduct our experiment in Pytorch (1.9.0+cuda10.2) using latest version of Python (3.8.10). Additional environment detail is provided in project\footnote{\url{https://www.dropbox.com/s/a6qioejg6yrxo7s/SLOF.tar.gz?dl=0}}.

Following are the list of configurations we used for our project:
\begin{itemize}
    \item Train/Val Batch Size: Ours(16/12 - 8Gpus), Finetune(8/12 - 4 Gpus)
    \item Number of iterations on RAFT: 12
    \item Loss: CosineSimilarity, Optical Flow L1-loss
    \item Optimizer: AdamW
    \item Scheduler: OneCycleLR
    \item EarlyStopping: Patience (5) and Min-delta (1e-4)
    \item Others: Gradient Clipping, GradScaler
    \item Dataset: FLOW360\footnote{\url{https://www.dropbox.com/s/nvzhazq99bg46f2/FLOW360_train_test.zip?dl=0}. Note that for better visualization please clip optical flow in the range of (-40,40) or lower.}
\end{itemize}

We suggest our readers to refer sample videos\footnote{\url{https://www.dropbox.com/s/54mmjvoz6844mci/trailer.mp4?dl=0}}$^,$\footnote{\url{https://www.dropbox.com/s/gvihrzj528d92uj/videos.zip?dl=0} We recommend to use VLC-Media player to play these videos.} for demo purpose.

\section{Impact of Rotational Invariance}
We perform additional experiments to quantify the impact of rotational invariance (see table below, Table.\ref{tab:rotinv}). The improvement is noticeable as seen in LiteFLowNet360~\cite{revisiting} and Tangent Images~\cite{tangent}-based optical flow estimation with rotational invariance.

\begin{table}[ht]
\scriptsize
\caption{Impact of rotational invariance}
\centering
\begin{tabular}{|c|c|c|}
\hline
Method & w/o Rot.Inv (EPE) & w Rot.Inv (EPE) \\
\hline
LiteFLowNet360~\cite{revisiting} & 3.95 & \textbf{2.52} \\
Tangent Images~\cite{tangent} & 3.57 & \textbf{1.78} \\
\hline
\end{tabular}
\label{tab:rotinv}
\end{table}

\begin{figure}[!t]
\centering
    \includegraphics[width=\textwidth]{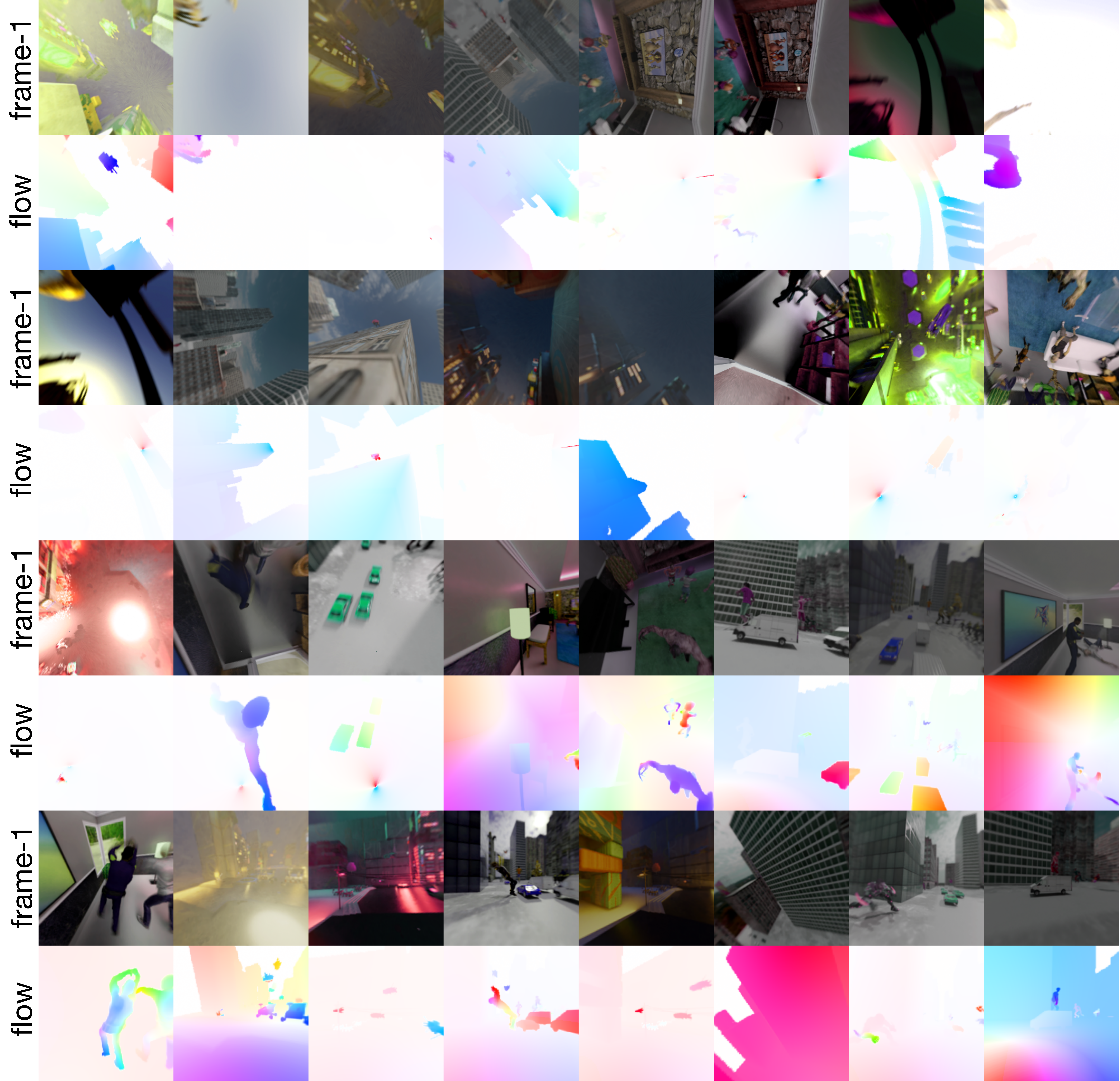}
    \caption{\textbf{More Motion and Scene Diversity - Train Set}. Illustrating motion and scene diversity via randomly sampled tangential plane from FLOW360 dataset. Flow360 contains range of scene complexity governing varied properties like texture, illumination, human, building, cars and other 3D assets. Similarly, the various level of motion complexity can be seen for similar scenes ranging from smaller to larger displacement. As explained in the main paper the dataset also contains other complexities like motion blur, camera focus/defocus, camera distortion and environmental effects.}
    \label{fig:trainmont}
\end{figure}

\begin{figure}[!t]
\centering
    \includegraphics[width=\textwidth]{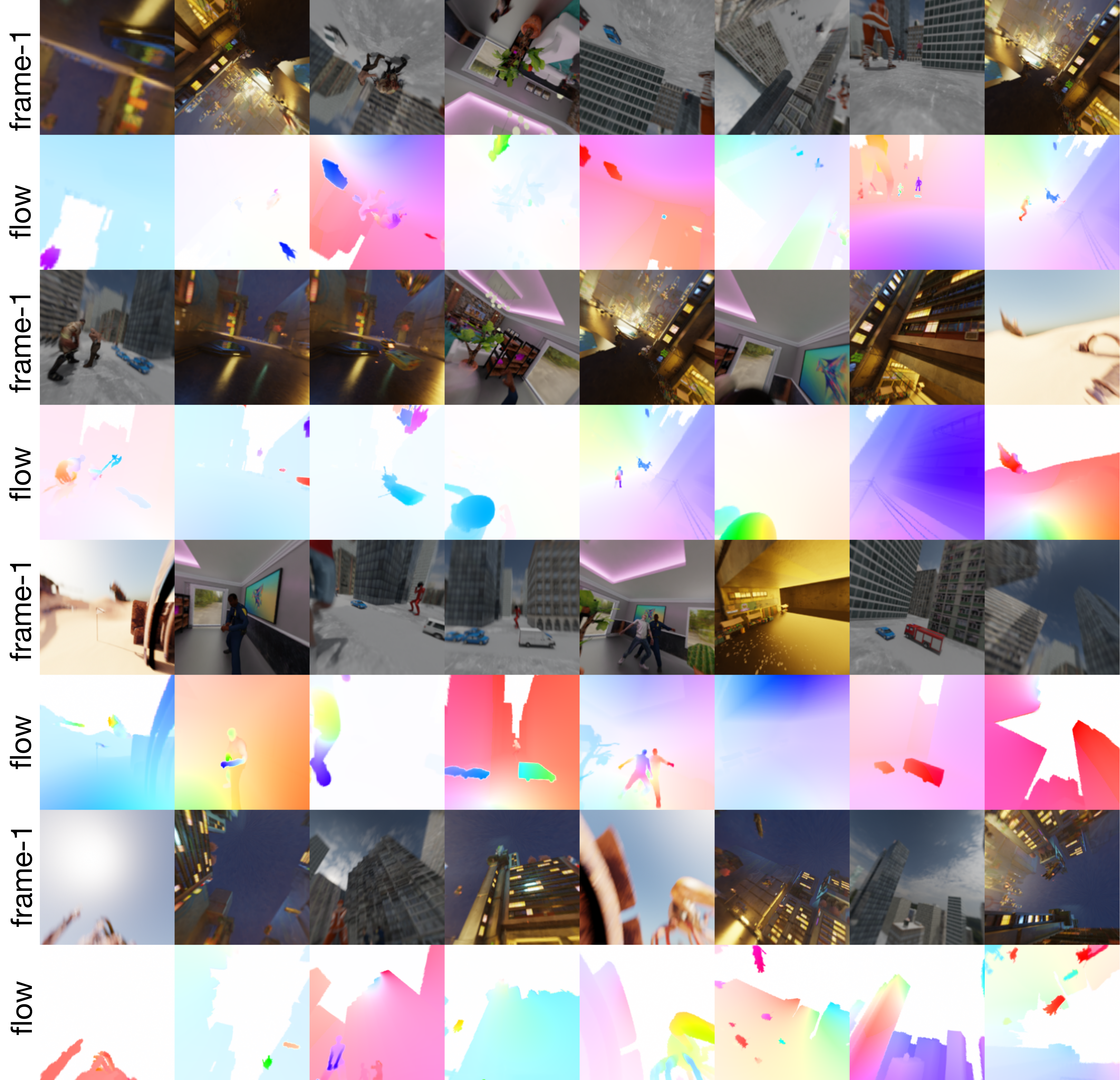}
    \caption{\textbf{More Motion and Scene Diversity - Test Set}. Illustrating motion and scene diversity via randomly sampled tangential plane from FLOW360 dataset. Flow360 contains range of scene complexity governing varied properties like texture, illumination, human, building, cars and other 3D assets. Similarly, the various level of motion complexity can be seen for similar scenes ranging from smaller to larger displacement. As explained in the main paper the dataset also contains other complexities like motion blur, camera focus/defocus, camera distortion and environmental effects.}
    \label{fig:testmont}
\end{figure}

	%
	%
\end{document}